\documentclass{article}

\usepackage{iclr2027_conference,times}
\usepackage{amsmath,amsfonts,bm}

\def\eqref#1{equation~\ref{#1}}
\def\1{\bm{1}}

\def\vtheta{{\bm{\theta}}}

\def\vb{{\bm{b}}}

\def\vd{{\bm{d}}}
\def\ve{{\bm{e}}}

\def\vg{{\bm{g}}}

\def\vp{{\bm{p}}}

\def\vr{{\bm{r}}}

\def\vv{{\bm{v}}}
\def\vw{{\bm{w}}}
\def\vx{{\bm{x}}}
\def\vy{{\bm{y}}}
\def\vz{{\bm{z}}}

\def\mA{{\bm{A}}}

\def\mI{{\bm{I}}}
\def\mJ{{\bm{J}}}

\DeclareMathAlphabet{\mathsfit}{\encodingdefault}{\sfdefault}{m}{sl}
\SetMathAlphabet{\mathsfit}{bold}{\encodingdefault}{\sfdefault}{bx}{n}

\usepackage{amsmath}
\usepackage{amssymb}
\usepackage{booktabs}
\usepackage{float}
\usepackage{graphicx}
\usepackage{xcolor}
\usepackage{tikz}
\usetikzlibrary{arrows.meta,calc,positioning}
\usepackage{enumitem}
\usepackage{hyperref}
\usepackage{url}

\definecolor{mthreeblue}{HTML}{DDEBF7}
\definecolor{mthreepurple}{HTML}{E8DDF5}
\definecolor{mthreeorange}{HTML}{FCE4D6}
\definecolor{mthreegray}{HTML}{EEEEEE}
\newcommand{\method}{AxisQ}

\newcommand{\vepsilon}{\boldsymbol{\epsilon}}

\title{\method{}: Depth-Adaptive\\
Channel--Frequency Quantization for\\
Camouflaged Object Detection}

\author{%
Chi Ruan$^1$, Tianqi Li$^2$, Xin He$^3$, Xue Geng$^4$, Ming-Ming Cheng$^2$, Yun Liu$^2$\\
\normalfont $^1$SpinEntropy, $^2$Nankai University, $^3$Tianjin University of Technology, $^4$I2R, A*STAR
}

\definecolor{publiclinkcolor}{rgb}{0,0,0.5}
\hypersetup{%
  colorlinks=true,
  linkcolor=publiclinkcolor,
  citecolor=publiclinkcolor,
  urlcolor=publiclinkcolor,
  pdftitle={AxisQ: Depth-Adaptive Channel-Frequency Quantization for Camouflaged Object Detection},
  pdfauthor={Chi Ruan, Tianqi Li, Xin He, Xue Geng, Ming-Ming Cheng, Yun Liu},
  pdfsubject={Public preprint}
}

\iclrfinalcopy

\begin{document}

\maketitle
% The template sets a conference-publication header inside \maketitle.
% Use the review-status header for this signed public version.
\lhead{Under review as a conference paper at ICLR 2027}

\begin{abstract}
Camouflaged object detection (COD) depends on weak texture and boundary cues
that can be destroyed by aggressive post-training quantization (PTQ).
Existing PTQ methods mainly optimize quantizers or
reconstruct network blocks, while using a single precision-allocation domain
throughout hierarchical backbones.
Instead, we introduce \method{}, a depth-adaptive bit-allocation framework motivated by
the observation that the useful allocation axis changes with feature depth.
Under a fixed per-layer average activation-bit budget, \method{} allocates
precision in the channel domain for shallow features and in the local-frequency
domain for deep features, using COD-aware sensitivity to better
preserve subtle foreground regions and ambiguous boundaries.
We also formulate a boundary-aware Hessian approximation for
reconstruction by incorporating boundary-weighted gradient and error statistics.
Experiments across multiple COD models and diverse benchmarks demonstrate
improved accuracy over the strongest baselines at low bit widths, with relative weighted F-measure gains of 12.7\% for CFRN and 33.3\% for CamoFormer-P on CAMO under W2A2 quantization. These results support adapting the allocation domain to feature depth
as a transferable quantization strategy for hierarchical COD models.
Code: \url{https://github.com/chi558/AxisQ}.
\end{abstract}

\section{Introduction}
\label{sec:introduction}

\begin{figure}[!t]
  \centering
  \includegraphics[width=\linewidth]{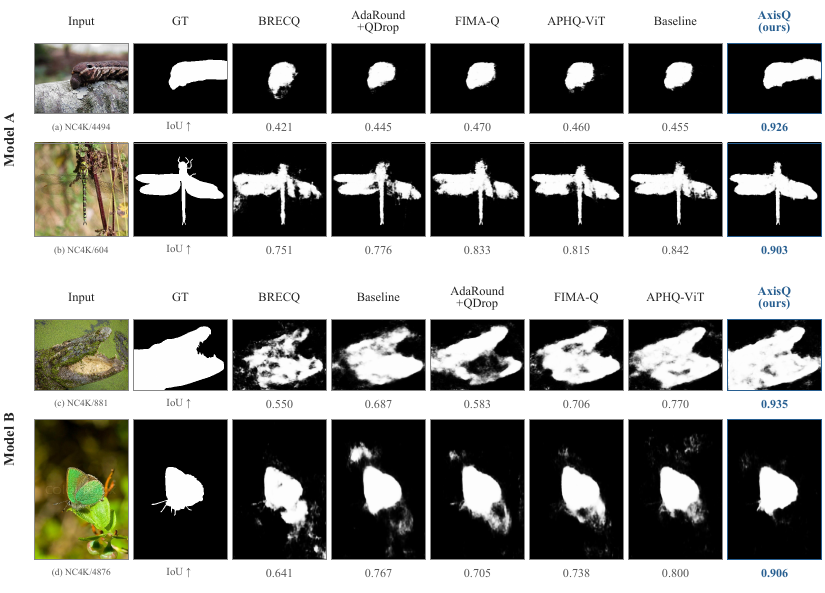}
  \caption{Qualitative comparison of quantization methods on two COD models
  at W2A2. Model A is CFRN, and Model B is CamoFormer-P.
  Each pair of rows corresponds to one model, with IoU reported for each
  prediction against the ground-truth mask.
  The compared methods show incomplete foreground regions and false detections
  in the background. In these examples, \method{} recovers more complete object
  shapes and clearer boundaries while reducing background false positives.}
  \label{fig:w2a2-qualitative}
\end{figure}

Camouflaged object detection (COD) segments objects whose color, texture, and
boundaries resemble their surroundings~\citep{fan2020cod}.
This makes accurate prediction unusually dependent on low-amplitude and
spatially structured evidence.
Modern state-of-the-art COD models~\citep{hu2023hitnet,yin2024camoformer,song2025cfrn} usually use hierarchical
Transformers to combine fine local details with semantic context, but their
memory and arithmetic cost complicate deployment.
Post-training quantization (PTQ) is attractive because it compresses a trained
model without end-to-end retraining~\citep{nagel2020adaround,li2021brecq}.
At low bit widths, however, a small calibration error can erase precisely the
cues that distinguish a camouflaged foreground from its background.
Figure~\ref{fig:w2a2-qualitative} illustrates foreground omissions and
inaccurate boundaries under low-bit quantization.

Sensitivity-aware mixed precision and curvature-based reconstruction
already account for the unequal effects of quantization
perturbations~\citep{li2021brecq,wu2025fimaq}.
In the context of COD, recent calibration work addresses the loss of weak
target cues caused by activation-range imbalance~\citep{li2026codtdq}.
However, these methods typically retain the same precision-allocation domain
throughout the model.
Whether this domain remains suitable as features change from shallow to deep
stages is less explored.
This motivates exploring whether allocating precision in different domains
across stages can improve quantization performance.

\begin{figure}[!t]
  \centering
  \includegraphics[width=\linewidth]{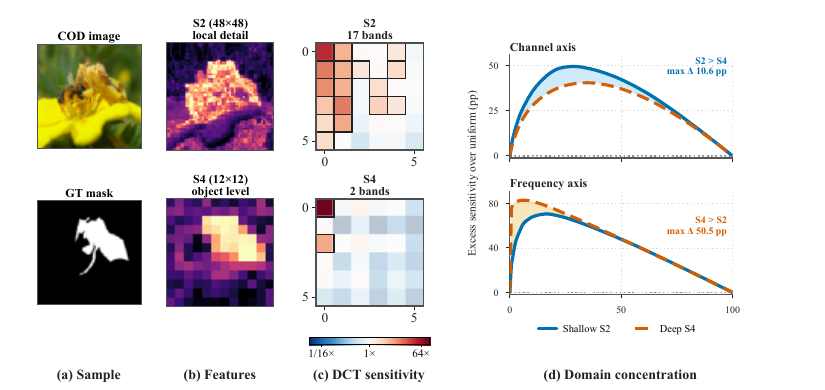}
  \caption{Depth-dependent sensitivity on a selected COD image.
  (a) shows the input and ground-truth mask, and (b) shows
  independently normalized S2 and S4 feature responses.
  In (c), 17 bands in S2 and two in S4 account for 80\% of DCT sensitivity.
  (d) plots cumulative sensitivity above uniform versus the
  percentage of top-ranked channels or frequency bands, with separate vertical scales.
  Sensitivity is more concentrated across channels in S2 and across frequency
  bands in S4, supporting channel-wise allocation in shallow stages and
  local-frequency allocation in deep stages.}
  \label{fig:motivation}
\end{figure}

To investigate this possibility, we examine how feature characteristics at
different depths affect the choice of precision-allocation domain.
Shallow features capture local patterns, and their channels differ in
sensitivity to quantization.
Allocating bits by channel allows more precision to be assigned to those
carrying weak target cues.
Deeper features combine information over larger spatial regions.
In these features, local frequency bands separate smooth spatial content
from variations associated with boundaries and textures, allowing these
components to receive different precision.
This difference suggests that shallow and deep features may benefit from
different allocation domains.
The analysis in Figure~\ref{fig:motivation} supports channel-wise allocation
in shallow stages and local-frequency allocation in deep stages.

\begin{figure}[H]
  \centering
  \includegraphics[width=\linewidth]{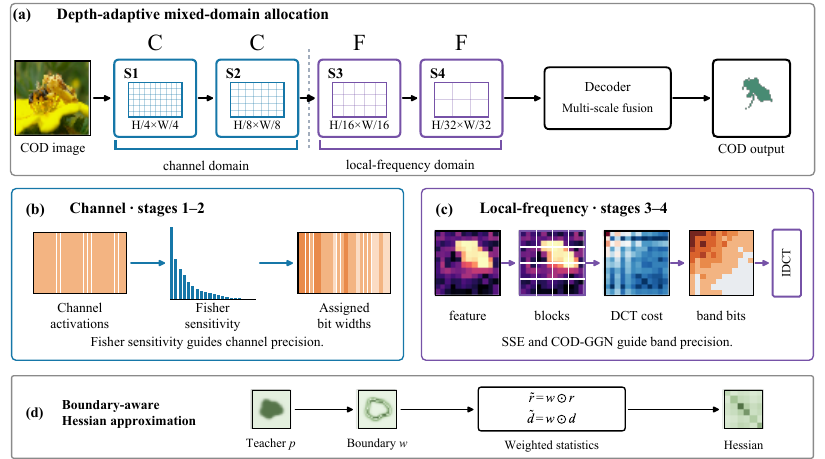}
  \caption{Overview of \method{} for a four-stage hierarchical encoder.
  Channel-domain allocation is used in shallow and eligible non-frequency
  paths, while deep encoder features use local-frequency allocation.  The
  strip and grid visualize channel-wise and frequency-wise activation-bit
  assignments at W2A2; darker orange indicates higher precision, and gray
  denotes zero-bit bands.  (d) sketches the auxiliary boundary-aware
  Hessian approximation using teacher-derived boundary weights and weighted
  gradient statistics.}
  \label{fig:overview}
\end{figure}

Accordingly, we propose \method{}, which allocates precision across channels in shallow stages
and across local-frequency bands in deep stages.
Figure~\ref{fig:overview} shows the overall framework.
Figure~\ref{fig:w2a2-qualitative} shows improved foreground and boundary preservation at W2A2 across both architectures.
Under a fixed per-layer average activation-bit budget, \method{} uses
COD-aware sensitivity to assign higher precision to components important for
foreground and boundary prediction.
We also formulate a boundary-aware Hessian approximation for reconstruction
using boundary-weighted gradient and error statistics.
Experiments demonstrate \method{}'s gains across COD models and precision settings. At W2A2, \method{} improves mean weighted F-measure over the best competing results by 6.6\% on CFRN and 21.5\% on CamoFormer-P across four benchmarks.

In summary, our main contributions include:
\begin{itemize}[leftmargin=*]
  \item We propose \method{}, a depth-adaptive mixed-domain PTQ framework allocating activation precision across shallow-stage channels and deep-stage local-frequency bands under a fixed per-layer average-bit budget, with an auxiliary boundary-aware Hessian approximation.
  \item We develop task-sensitive channel-bit allocation for COD. Label-assisted empirical-Fisher costs measure channel sensitivity, while reverse water filling followed by exact budget correction assigns channel bit widths while satisfying each layer's budget exactly.
  \item We derive a positive-semidefinite COD-GGN cost for local-frequency
  allocation. It balances foreground and background contributions and
  emphasizes multi-scale boundaries and uncertain pixels, using one
  first-order probe per layer and calibration batch.
\end{itemize}

\section{Related Work}
\label{sec:related}

\paragraph{Camouflaged object detection.}
COD methods aim to distinguish concealed targets from similar backgrounds
while recovering fine object structures~\citep{mei2021pfnet,sun2022bgnet}.
Early approaches addressed target localization through classification-assisted
segmentation in ANet~\citep{le2019anabranch} and a search-and-identification
framework in SINet~\citep{fan2020cod}.
Later work introduced complementary information to recover less visible
regions: ZoomNet fused features across image scales~\citep{pang2022zoomnet},
while FDCOD incorporated frequency-domain cues~\citep{zhong2022fdcod}.
Transformer-based methods combine semantic context and spatial detail.
HitNet used high-resolution feedback to refine low-resolution
representations~\citep{hu2023hitnet}, and CamoFormer progressively refined
predictions through masked separable attention~\citep{yin2024camoformer}.
Our work examines how to preserve foreground localization and detailed
structures under low-bit quantization.

\paragraph{Post-training quantization.}
PTQ research addresses quantization errors through reconstruction and adapts
quantizers to feature distributions through calibration~\citep{banner2019post,liu2023noisyquant}.
AdaRound and BRECQ match full-precision outputs by optimizing rounding at
layer and block levels,
respectively~\citep{nagel2020adaround,li2021brecq}, while QDrop improves
robustness to activation quantization~\citep{wei2022qdrop}.
Vision Transformers require calibration tailored to attention and
normalization layers~\citep{ding2022apqvit}.
This motivates attention-ranking constraints~\citep{liu2021vitptq},
distribution-aware quantizers in PTQ4ViT and FQ-ViT~\citep{yuan2022ptq4vit,lin2022fqvit},
and scale reparameterization in RepQ-ViT~\citep{li2023repqvit}.
These methods quantize existing feature representations; our work explores
depth-dependent allocation domains.

\paragraph{Sensitivity-aware quantization.}
Quantization errors affect model predictions unequally~\citep{dong2020hawqv2,tai2024mptqvit}.
HAWQ uses Hessian information to allocate layer-wise precision~\citep{dong2019hawq}.
For reconstruction, FIMA-Q weights feature errors with diagonal and low-rank
Fisher approximations~\citep{wu2025fimaq}, while LS-ViT fits curvature through
least squares~\citep{hwang2026lsvit}.
COD-TDQ constrains token-group ranges to protect weak boundary cues from
background-dominated quantization~\citep{li2026codtdq}.
These methods consider information importance beyond numerical error,
motivating allocation domains that expose task-relevant features at different depths.

\section{Method}
\label{sec:method}

This section presents our depth-adaptive mixed-domain quantization framework
for hierarchical COD models.  It covers channel-wise allocation in shallow
stages, frequency-wise allocation in deep stages, and the corresponding
reconstruction procedure.

\subsection{Depth-adaptive mixed-domain allocation}
\label{sec:framework}

Low-bit quantization can suppress the weak foreground and boundary cues
required for COD.  Figure~\ref{fig:motivation} suggests that sensitivity is
more concentrated across channels in shallow features and across
local-frequency bands in deep features, so we adapt the allocation domain to
network depth.

Let $f(\vx;\vtheta)$ denote a trained hierarchical COD model, and let
$\mA_\ell\in\mathbb{R}^{B\times N_\ell\times C_\ell}$ denote the input
activation of layer $\ell$.  We use an asymmetric $b$-bit uniform quantizer:
\begin{equation}
  Q_b(a;s,z)=s\left[
  \operatorname{clip}\!\left(\operatorname{round}(a/s)+z,0,2^b-1\right)-z
  \right],
  \label{eq:quantizer}
\end{equation}
where $s$ and $z$ are the quantization scale and zero point, respectively.
Weights are quantized at a fixed precision of $W_b$ bits.  The activation of
layer $\ell$ is partitioned into $K_\ell$ allocation units, namely channels or
local-frequency bands, and each unit $k$ receives a bit width $b_{\ell,k}$
from a set of allowed tiers $\mathcal{B}$.  We seek
\begin{equation}
  \min_{\{b_{\ell,k}\}\subset\mathcal{B}}\;
  \sum_{k=1}^{K_\ell}\mathcal{D}_{\ell,k}(b_{\ell,k})
  \quad\text{s.t.}\quad
  \frac{1}{K_\ell}\sum_{k=1}^{K_\ell}b_{\ell,k}=A_b,
  \label{eq:allocation}
\end{equation}
where $\mathcal{D}_{\ell,k}(b)$ estimates how quantizing unit $k$ at $b$ bits
affects COD predictions.

As shown in Figure~\ref{fig:overview}, \method{} instantiates
Eq.~(\ref{eq:allocation}) with two depth-specific branches followed by a
shared reconstruction stage.  Stages 1--2 use channels as units with
label-assisted empirical-Fisher costs, while stages 3--4 use block-DCT
local-frequency bands with costs that combine reconstruction error and
COD-GGN sensitivity.  The stage partition is fixed before calibration and
retained across precision settings; frequency allocation takes precedence
when both branches are applicable.  After calibration, the assignments are
frozen for blockwise reconstruction, where the boundary-aware Hessian
approximation serves as an auxiliary extension.
Sections~\ref{sec:shallow}--\ref{sec:boundary-reconstruction} develop these
three components in order.

\subsection{Task-sensitive channel-bit allocation}
\label{sec:shallow}

\paragraph{Empirical-Fisher channel cost.}
We compute the supervised binary-cross-entropy loss on a small paired COD
calibration subset.  For layer $\ell$, channel $c$, and candidate bit tier
$b$, the per-channel empirical-Fisher distortion~\citep{kunstner2019limitations} is
\begin{equation}
  D_{\ell,c}^{(b)}=\mathbb{E}_{(\vx,\vy)}\!\left[
  \left(\sum_n
  \frac{\partial\mathcal{L}}{\partial\mA_{\ell,n,c}}
  \left(Q_b(\mA_{\ell,n,c})-\mA_{\ell,n,c}\right)\right)^2\right],
  \label{eq:fisher-channel}
\end{equation}
where $n$ indexes spatial or token positions and any remaining non-channel
dimensions, $Q_b$ is the $b$-bit activation quantizer, and the expectation is
taken over image-label pairs in the calibration subset.  The inner sum
approximates the first-order loss change caused by quantizing channel $c$;
squaring and averaging this response gives the channel distortion used for bit
allocation.  The score retains interactions within a channel while ignoring
cross-channel interactions.  It uses only first derivatives and 64
ground-truth masks.

\paragraph{Reverse water filling.}
In Eq.~(\ref{eq:allocation}), $\mathcal{D}_{\ell,c}(b)=D_{\ell,c}^{(b)}$.
We use the reference-tier costs in a budget-preserving heuristic.
Given a target average activation precision $A_b$, we initialize a continuous
precision for each channel by adjusting $A_b$ according to its log sensitivity
relative to the layer mean:
\begin{equation}
 \widetilde b_{\ell,c}
 =A_b+\gamma\left[
 \log_2\!\left(D_{\ell,c}^{(b_{\mathrm{ref}})}+\epsilon\right)
 -\frac{1}{C_\ell}\sum_{j=1}^{C_\ell}
 \log_2\!\left(D_{\ell,j}^{(b_{\mathrm{ref}})}+\epsilon\right)\right],
 \label{eq:waterfill}
\end{equation}
where $b_{\mathrm{ref}}$ is a reference bit tier, $\gamma$ controls the
strength of sensitivity-based reallocation, and $\epsilon$ improves numerical
stability.  Channels with larger distortion therefore receive higher initial
precision.  We round the continuous values to the allowed bit tiers and
greedily adjust the assignments until
$C_\ell^{-1}\sum_c b_{\ell,c}=A_b$ exactly.
This layer-wise allocation gives more bits to sensitive channels while
preserving the target average precision $A_b$ for each layer.

\subsection{COD-GGN task-aware frequency allocation}
\label{sec:codggn}
\label{sec:deep}

\paragraph{Block-frequency representation and reconstruction cost.}
For deep activations, we partition the spatial feature map into local blocks
and apply an orthonormal 2-D DCT to each block.  Coefficients at the same
within-block frequency coordinate are grouped into a frequency band and share
one bit width.  Let $\mathcal{T}$ denote the block-DCT transform and
$\vb_\ell$ the band-wise bit assignment.  The quantized activation is
reconstructed as
\begin{equation}
 \widehat{\mA}_\ell
 =\mathcal{T}^{-1}\!\left(
 Q_{\vb_\ell}\left(\mathcal{T}(\mA_\ell)\right)
 \right),
 \label{eq:dct-sandwich}
\end{equation}
where $\mathcal{T}$ is the block-DCT transform.

For frequency band $k$ and candidate bit tier $b$, let $u_{i,k}$ denote the
DCT coefficient from calibration sample $i$.  We estimate the reconstruction
cost by
\begin{equation}
 C_{k,b}^{\mathrm{SSE}}
 =\min_r\sum_i
 \left(Q_{b,r}(u_{i,k})-u_{i,k}\right)^2 .
 \label{eq:freq-sse}
\end{equation}
where $r$ indexes candidate clipping ranges.  The minimum over $r$ selects the
clipping range that gives the smallest reconstruction error for each band and
bit tier.  This SSE term measures
coefficient distortion without task weighting; the COD-aware cost introduced
below accounts for the task sensitivity of different frequency bands.  These
SSE values provide a reconstruction-only reference for comparing candidate bit
tiers before task sensitivity is incorporated.  Each band is evaluated
independently, so candidate tiers share a common reconstruction criterion.

\paragraph{COD-aware curvature and frequency cost.}
Reconstruction error weights all DCT coefficient errors equally, although COD is
more sensitive to errors near object boundaries, uncertain predictions, and
small foreground regions.  From the full-precision teacher probability map
$\vp=\sigma(\vz)$, we define the output-space task metric
\begin{equation}
\begin{gathered}
 \Omega=\operatorname{Norm}\!\left(R(\vp)[1+\alpha B(\vp)]U(\vp)\right),\\
 B(\vp)=\frac{1}{|\mathcal{Q}|}\sum_{q\in\mathcal{Q}}
 \left|\operatorname{AvgPool}_q(\vp)-\vp\right|,\\
 U(\vp)=\eta+(1-\eta)\,4\vp(1-\vp).
\end{gathered}
\label{eq:omega}
\end{equation}
Here $R$ balances foreground and background mass, $B$ captures multi-scale
boundary variation, and $U$ emphasizes uncertain predictions while retaining a
nonzero floor.  Together, these factors separate global mass balance from local
task structure: $R$ prevents large regions from dominating, while $B$ and $U$
highlight boundary changes and uncertain predictions.  Their product is
normalized independently for each image, so the probe represents relative task
importance rather than output magnitude.

Using a Rademacher probe, we map this output-space metric to the activation at
layer $\ell$ through a vector--Jacobian product (VJP).  This produces a
positive-semidefinite COD-aware generalized Gauss--Newton (COD-GGN) curvature surrogate without constructing a
full Hessian~\citep{martens2020natural}.  The probe response and the quantization error are expressed in
the same block-DCT basis.  For frequency band $k$ and candidate bit tier $b$,
we define the task-aware cost and its fusion with the reconstruction cost as
\begin{equation}
\begin{gathered}
 C_{k,b}^{\mathrm{task}}=\sum_i (g_{i,k})^2(\ve_{i,k}^{(b)})^2,\\
 C_{k,b}=(1-\beta)\frac{C_{k,b}^{\mathrm{SSE}}}
 {\sum_j C_{j,b_{\mathrm{ref}}}^{\mathrm{SSE}}}
 +\beta\frac{C_{k,b}^{\mathrm{task}}}
 {\sum_j C_{j,b_{\mathrm{ref}}}^{\mathrm{task}}}.
\end{gathered}
\label{eq:task-cost}
\end{equation}
Here $g_{i,k}$ and $\ve_{i,k}^{(b)}$ are the transformed probe response and
quantization error for sample $i$ and band $k$.  A larger probe response makes
the same quantization error more costly and therefore favors a higher bit tier.
Reference-tier normalization keeps the two terms comparable, while $\beta$
controls the influence of task sensitivity.  With
$\mathcal{D}_{\ell,k}(b)=C_{k,b}$ in Eq.~(\ref{eq:allocation}), the
Lagrangian relaxation gives $b_k=\arg\min_b(C_{k,b}+\lambda b)$.
We then greedily correct the assignments so that their average equals $A_b$ exactly.

The SSE term measures coefficient distortion, whereas the task term increases
the penalty for errors in COD-sensitive directions.  Setting $\beta=0$ recovers
a reconstruction-only frequency allocator; larger $\beta$ shifts the allocation
toward task-sensitive bands.  After correction, the frozen band assignments are
passed to the shared reconstruction procedure below.  All task costs are
computed during calibration and discarded after the assignments are frozen, so
the selected quantizers introduce no additional inference-time COD computation.

\subsection{Boundary-aware reconstruction}
\label{sec:boundary-reconstruction}

After allocations are frozen, we reconstruct the model block by block using the
existing Hessian-based reconstruction objective~\citep{hwang2026lsvit}, AdaRound weight
variables~\citep{nagel2020adaround}, and QDrop~\citep{wei2022qdrop}.  This
pipeline is shared by matched baselines and \method{}; COD-GGN affects only
frequency calibration, while the boundary-aware extension is evaluated
separately.

To make reconstruction boundary-aware, we reweight the gradient and error
statistics before the existing Hessian fitting procedure.  For calibration
image $i$ at block $\ell$, let $\vr_{\ell,i}$ be the absolute teacher-loss
gradient and $\vd_{\ell,i}$ its absolute deviation from the full-precision
output.  With the teacher probability map $\vp_{\ell,i}$ resized to the block
grid, we use
\begin{equation}
\begin{aligned}
 \vw_{\ell,i}&=1+\alpha_{\mathrm{rec}}
 \left|\operatorname{AvgPool}_{k_\ell}(\vp_{\ell,i})-\vp_{\ell,i}\right|,\\
 \widetilde\vr_{\ell,i}&=\vw_{\ell,i}\odot\vr_{\ell,i},\qquad
 \widetilde\vd_{\ell,i}=\vw_{\ell,i}\odot\vd_{\ell,i}.
\end{aligned}
\label{eq:reconstruction-boundary-weight}
\end{equation}
The weight map is expanded to the flattened block statistics.  Larger local
prediction variation therefore gives boundary-sensitive responses more weight
in the curvature fit.  Applying the same map to the gradient and error vectors
preserves their correspondence with the original reconstruction statistics.
AdaRound, QDrop, and final-logit BCE reconstruction remain unchanged.  Because
the weights come from the teacher prediction, no additional segmentation labels
are required.

\section{Experiments}
\label{sec:experiments}

This section evaluates \method{} across various COD models and benchmarks under
different quantization settings. We describe the experimental setup, present
comparisons with existing methods, and examine the contributions of individual
components.

\subsection{Experimental setup}

\paragraph{Models and quantization settings.}
We quantize the Swin-based CFRN model~\citep{song2025cfrn,liu2021swin} and
CamoFormer-P~\citep{yin2024camoformer,wang2022pvtv2} at input resolution $384\times384$.
Models A and B in Figure~\ref{fig:w2a2-qualitative} denote CFRN and CamoFormer-P,
respectively.
The main comparison evaluates both models at W2A2 and W4A4, with additional
W2A3, W2A4, and W3A3 results on CFRN.
These settings allow us to evaluate \method{}'s performance across different
precision levels.
Patch-embedding activations remain at A8 for every method and precision setting.

\paragraph{Datasets and evaluation metrics.}
Evaluation uses the standard CAMO, CHAMELEON, COD10K, and NC4K test
sets~\citep{le2019anabranch,skurowski2018chameleon,fan2020cod,lv2021nc4k}.
We report S-measure $S_\alpha$~\citep{fan2017smeasure} for structural similarity
and weighted F-measure $F_\beta^w$~\citep{margolin2014foreground} for spatially
weighted precision and recall. Mean E-measure $E_\phi$~\citep{fan2018emeasure}
assesses local and global alignment, averaged across thresholds.
Maximum F-measure $F_\beta^m$ is the highest F-score across thresholds,
while MAE measures the mean absolute pixel-wise prediction error.
Higher is better except for MAE.
Macro metrics give equal weight to each dataset. Macro quality averages
$S_\alpha$, $F_\beta^w$, $E_\phi$, $F_\beta^m$, and $1-\mathrm{MAE}$
with equal weights across metrics and datasets.

\paragraph{Implementation details.}
All calibration samples are drawn from the training set.
The calibration pool contains 128 images, processed with batch size 4.
We use 64 paired image--mask samples with batch size 2 to estimate
channel-domain Fisher costs, and 16 images to estimate deep frequency and
COD-GGN costs. Block reconstruction uses 1,024 images, an effective batch
size of 32, and 20,000 optimization iterations per block.
QDrop is applied with probability 0.5.
We set the channel allocation spread to $\gamma=0.30$ and use
$12\times12$ DCT blocks with eight candidate clipping ratios.
For COD-GGN, we use pooling sizes $\mathcal{Q}=\{3,7,15\}$, boundary weight
$\alpha=2$, uncertainty floor $\eta=0.1$, and cost mixing weight $\beta=0.10$.
The output weights are normalized to mean one per image and capped at 8.
Both frequency cost terms are normalized at the reference tier
$b_{\mathrm{ref}}=2$. These settings, the stage split, and the reconstruction
length remain fixed across precision budgets.

\paragraph{Compared methods.}
We compare with LS-ViT~\citep{hwang2026lsvit}, BRECQ~\citep{li2021brecq},
AdaRound+QDrop~\citep{nagel2020adaround,wei2022qdrop},
FIMA-Q~\citep{wu2025fimaq}, and APHQ-ViT~\citep{wu2025aphqvit} on both
models, and additionally with RepQ-ViT~\citep{li2023repqvit} and
COD-TDQ~\citep{li2026codtdq} on CFRN.
These methods cover calibration-based quantization, adaptive rounding, and
curvature-based reconstruction.
All comparator results use local COD adaptations.
Reconstruction-based methods use 20,000 optimization iterations per block;
RepQ-ViT and COD-TDQ use no reconstruction.
COD-TDQ retains full-precision attention matrix
multiplications; its W2A2 row is a local four-level affine extension,
not an author-released W2A2 result.
FP32 results provide the full-precision reference for each model.

\subsection{Quantitative results}
\label{sec:quantitative-results}

\method{} delivers substantial gains under W2A2 quantization across both
backbones, as shown in Table~\ref{tab:main-comparison}.
It leads all 20 dataset--metric comparisons on CFRN and 19 on CamoFormer-P.
On CAMO, its weighted F-measure exceeds the best competing result by $0.082$
on CFRN and $0.125$ on CamoFormer-P.
The consistent improvements across models and evaluation measures support the
effectiveness of \method{} in preserving segmentation quality under a tight
bit budget.

% CFRN data and boldface: d2cdd22, tables/table1_w2a2.tex and table2_w4a4.tex.
% CamoFormer-P data and boldface: 41544d2, tables/table1_main_comparison.tex.
% CamoFormer-P W2A2 and W4A4 source metrics: JSON records in data/ (2026-09-13).
\begin{table}[t]
\caption{Quantitative comparison on CFRN and CamoFormer-P across four COD
benchmarks at W4A4 and W2A2. FP32 denotes the full-precision reference.
Bold indicates the best quantized results within each model and precision setting.}
\label{tab:main-comparison}
\label{tab:w2a2}
\label{tab:w4a4-sota}
\centering
\scriptsize
\setlength{\tabcolsep}{1.5pt}
\resizebox{\linewidth}{!}{%
\begin{tabular}{@{}l*{4}{ccccc}@{}}
\toprule
& \multicolumn{5}{c}{CAMO}
& \multicolumn{5}{c}{CHAMELEON}
& \multicolumn{5}{c}{COD10K}
& \multicolumn{5}{c}{NC4K}\\
\cmidrule(lr){2-6}\cmidrule(lr){7-11}
\cmidrule(lr){12-16}\cmidrule(lr){17-21}
Method
& $S_\alpha$$\uparrow$ & $F_\beta^w$$\uparrow$ & $E_\phi$$\uparrow$ & $F_\beta^m$$\uparrow$ & MAE$\downarrow$
& $S_\alpha$$\uparrow$ & $F_\beta^w$$\uparrow$ & $E_\phi$$\uparrow$ & $F_\beta^m$$\uparrow$ & MAE$\downarrow$
& $S_\alpha$$\uparrow$ & $F_\beta^w$$\uparrow$ & $E_\phi$$\uparrow$ & $F_\beta^m$$\uparrow$ & MAE$\downarrow$
& $S_\alpha$$\uparrow$ & $F_\beta^w$$\uparrow$ & $E_\phi$$\uparrow$ & $F_\beta^m$$\uparrow$ & MAE$\downarrow$\\
\midrule
\multicolumn{21}{c}{\textbf{CFRN}}\\
\midrule
FP32
& .876 & .843 & .932 & .877 & .042
& .912 & .874 & .959 & .898 & .020
& .870 & .794 & .937 & .835 & .022
& .888 & .849 & .939 & .880 & .030\\
\midrule
\multicolumn{21}{c}{\textbf{W4A4}}\\
\midrule
RepQ-ViT
& .676 & .608 & .750 & .656 & .099
& .701 & .622 & .774 & .672 & .067
& .676 & .549 & .779 & .599 & .062
& .718 & .651 & .783 & .691 & .072\\
COD-TDQ
& .813 & .723 & .864 & .795 & .070
& .861 & .759 & .906 & .825 & .040
& .802 & .650 & .864 & .737 & .038
& .836 & .746 & .883 & .816 & .052\\
LS-ViT
& .860 & .816 & .916 & .855 & .048
& .906 & .857 & .951 & .887 & .023
& .850 & .754 & .915 & .802 & .026
& .875 & .824 & .925 & .862 & .035\\
BRECQ
& .864 & .823 & .919 & .861 & .047
& .907 & .857 & \textbf{.956} & .883 & .022
& .859 & .769 & .923 & .815 & .024
& .881 & .834 & .930 & .870 & .033\\
AdaRound+QDrop
& .863 & .823 & .917 & .861 & .047
& .907 & .860 & .951 & .886 & .022
& .859 & .771 & .923 & .815 & .024
& .881 & .836 & .929 & .871 & .033\\
FIMA-Q
& .865 & .825 & .920 & .863 & .046
& .907 & .861 & .951 & .888 & .022
& .859 & .772 & .923 & .816 & .024
& .881 & .836 & .929 & .871 & .033\\
APHQ-ViT
& .866 & .827 & .921 & .864 & .046
& .907 & .859 & .951 & .886 & .022
& .859 & .772 & .923 & .817 & \textbf{.024}
& .882 & .837 & .930 & .872 & \textbf{.033}\\
\method{} (ours)
& \textbf{.869} & \textbf{.832} & \textbf{.924} & \textbf{.870} & \textbf{.044}
& \textbf{.909} & \textbf{.865} & \textbf{.956} & \textbf{.893} & \textbf{.021}
& \textbf{.861} & \textbf{.775} & \textbf{.925} & \textbf{.820} & \textbf{.024}
& \textbf{.883} & \textbf{.838} & \textbf{.932} & \textbf{.873} & \textbf{.033}\\
\midrule
\multicolumn{21}{c}{\textbf{W2A2}}\\
\midrule
COD-TDQ (ext.)
& .261 & .162 & .244 & .218 & .614
& .239 & .133 & .236 & .171 & .662
& .249 & .082 & .234 & .112 & .648
& .258 & .138 & .240 & .185 & .629\\
RepQ-ViT
& .386 & .106 & .485 & .218 & .247
& .400 & .085 & .511 & .171 & .215
& .421 & .061 & .534 & .111 & .175
& .399 & .093 & .503 & .185 & .224\\
BRECQ
& .702 & .559 & .764 & .641 & .113
& .779 & .625 & .839 & .709 & .066
& .734 & .521 & .803 & .607 & .059
& .787 & .659 & .841 & .736 & .072\\
AdaRound+QDrop
& .713 & .583 & .749 & .668 & .102
& .803 & .681 & .859 & .750 & .052
& .759 & .580 & .806 & .654 & .045
& .805 & .700 & .845 & .767 & .060\\
FIMA-Q
& .729 & .612 & .767 & .694 & .097
& .804 & .686 & .851 & .756 & .048
& .766 & .596 & .819 & .668 & .044
& .814 & .716 & .856 & .779 & .057\\
APHQ-ViT
& .743 & .631 & .785 & .707 & .093
& .823 & .712 & .874 & .776 & .047
& .774 & .608 & .828 & .679 & .042
& .818 & .723 & .861 & .785 & .056\\
LS-ViT
& .746 & .645 & .794 & .725 & .089
& .826 & .722 & .873 & .786 & .045
& .782 & .624 & .836 & .693 & .040
& .824 & .735 & .867 & .794 & .054\\
\method{} (ours)
& \textbf{.798} & \textbf{.727} & \textbf{.861} & \textbf{.787} & \textbf{.071}
& \textbf{.851} & \textbf{.766} & \textbf{.924} & \textbf{.809} & \textbf{.036}
& \textbf{.793} & \textbf{.654} & \textbf{.862} & \textbf{.713} & \textbf{.036}
& \textbf{.834} & \textbf{.760} & \textbf{.890} & \textbf{.810} & \textbf{.048}\\
\midrule[0.8pt]
\multicolumn{21}{c}{\textbf{CamoFormer-P}}\\
\midrule
FP32
& .873 & .827 & .926 & .868 & .046
& .898 & .845 & .941 & .884 & .026
& .860 & .770 & .920 & .814 & .025
& .887 & .837 & .931 & .872 & .032\\
\midrule
\multicolumn{21}{c}{\textbf{W4A4}}\\
\midrule
LS-ViT
& .861 & .810 & .914 & .853 & .050
& .892 & .833 & .935 & .876 & .029
& .850 & .748 & .909 & .798 & .027
& .879 & .822 & .923 & .862 & .035\\
BRECQ
& .858 & .805 & .911 & .849 & .051
& .889 & .828 & .935 & .870 & .029
& .846 & .742 & .906 & .790 & .028
& .877 & .820 & .922 & .859 & .035\\
AdaRound+QDrop
& .863 & .813 & .914 & .857 & .049
& .891 & .833 & .935 & .876 & .028
& .849 & .748 & .908 & .797 & .027
& .879 & .823 & .923 & .862 & \textbf{.034}\\
FIMA-Q
& .863 & .812 & .914 & .857 & .050
& .892 & .833 & .937 & .876 & .028
& .849 & .748 & .909 & .797 & .027
& .879 & .822 & .923 & .862 & .035\\
APHQ-ViT
& .862 & .810 & .914 & .853 & .049
& .891 & .833 & .937 & \textbf{.876} & \textbf{.028}
& .850 & .750 & .908 & .798 & \textbf{.027}
& .880 & .823 & .924 & .862 & .035\\
\method{} (ours)
& \textbf{.867} & \textbf{.817} & \textbf{.921} & \textbf{.858} & \textbf{.047}
& \textbf{.893} & \textbf{.834} & \textbf{.938} & \textbf{.876} & \textbf{.028}
& \textbf{.851} & \textbf{.751} & \textbf{.913} & \textbf{.800} & \textbf{.027}
& \textbf{.881} & \textbf{.825} & \textbf{.925} & \textbf{.864} & \textbf{.034}\\
\midrule
\multicolumn{21}{c}{\textbf{W2A2}}\\
\midrule
BRECQ
& .574 & .350 & .654 & .462 & .177
& .606 & .354 & .659 & .464 & .136
& .606 & .302 & .683 & .408 & .117
& .653 & .425 & .714 & .542 & .137\\
AdaRound+QDrop
& .576 & .347 & .608 & .456 & .164
& .630 & .398 & .652 & .513 & .108
& .645 & .367 & .712 & .468 & \textbf{.092}
& .698 & .499 & .745 & .602 & .110\\
FIMA-Q
& .581 & .362 & .629 & .462 & .167
& .642 & .412 & .692 & .507 & .110
& .640 & .363 & .709 & .468 & .102
& .698 & .501 & .751 & .607 & .116\\
APHQ-ViT
& .593 & .371 & .635 & .469 & .163
& .637 & .404 & .690 & .499 & .113
& .645 & .368 & .711 & .480 & .100
& .704 & .505 & .755 & .617 & .114\\
LS-ViT
& .598 & .375 & .631 & .484 & .179
& .636 & .394 & .671 & .507 & .134
& .632 & .340 & .687 & .470 & .122
& .696 & .479 & .736 & .610 & .131\\
\method{} (ours)
& \textbf{.671} & \textbf{.500} & \textbf{.727} & \textbf{.596} & \textbf{.146}
& \textbf{.716} & \textbf{.524} & \textbf{.771} & \textbf{.625} & \textbf{.104}
& \textbf{.671} & \textbf{.419} & \textbf{.734} & \textbf{.539} & .101
& \textbf{.735} & \textbf{.560} & \textbf{.788} & \textbf{.675} & \textbf{.106}\\
\bottomrule
\end{tabular}%
}
\end{table}

The advantage remains consistent at W4A4. At the reported three-decimal
precision, \method{} achieves the best quantized result in 33 of the 40
dataset--metric comparisons and ties for the best in the remaining seven.
This shows that its benefit extends to higher-precision quantization, where
reconstruction-based methods already perform closely.
Although the margins are smaller than at W2A2, \method{} maintains a
consistently strong position across both models and all four datasets.

The two backbones show different sensitivity to low-bit quantization.
Their FP32 mean weighted F-measures are relatively close, but CamoFormer-P
suffers a larger drop under the W2A2 LS-ViT baseline.
\method{} raises mean weighted F-measure over this baseline by approximately
$0.104$ on CamoFormer-P, compared with $0.045$ on CFRN.
This larger recovery on the more affected backbone further demonstrates \method{}'s ability to alleviate severe quantization degradation across
models.

Table~\ref{tab:precision} further demonstrates the applicability of \method{}
across different precision budgets on CFRN.
With shared allocation hyperparameters, it improves 94 of the 100
dataset--metric comparisons against precision-matched baselines, including
all 20 comparisons at W2A2, W3A3, and W4A4.
With weights fixed at two bits, mean weighted F-measure gains remain positive
as activation precision increases from two to four bits.
These results support using the same allocation strategy across different
combinations of weight and activation precision.

\subsection{Ablation studies and allocation analysis}
\label{sec:ablation-studies}
\label{sec:boundary-study}

Table~\ref{tab:ablation} suggests that the allocation domain should be
chosen according to feature depth.
The mixed-domain no-GGN variant improves macro weighted F-measure by
$0.025$ over the stronger frequency-only control, exceeding that control's
$0.013$ gain over the baseline.
This gap is consistent with the complementary roles of channel sensitivity
in shallow features and frequency structure in deep features.
The accompanying reduction in MAE supports an overall improvement in
prediction quality.

\begin{table}[!t]
\centering
\begin{minipage}[t]{0.47\linewidth}
\vspace{0pt}
\caption{\method{} gains across precisions.}
\label{tab:precision}
\centering
\footnotesize
\setlength{\tabcolsep}{1.2pt}
\begin{tabular}{lrrrrrr}
\toprule
Setting & $\Delta S_\alpha$ & $\Delta F_\beta^w$ &
$\Delta E_\phi$ & $\Delta F_\beta^m$ & $\Delta$MAE & Wins\\
\midrule
W2A2 & +.025 & +.045 & +.042 & +.030 & -.009 & 20/20\\
W2A3 & +.006 & +.009 & +.008 & +.008 & -.002 & 18/20\\
W2A4 & +.002 & +.004 & +.001 & +.005 & .000 & 16/20\\
W3A3 & +.006 & +.011 & +.010 & +.008 & -.002 & 20/20\\
W4A4 & +.008 & +.015 & +.008 & +.012 & -.003 & 20/20\\
\bottomrule
\end{tabular}
\end{minipage}\hfill
\begin{minipage}[t]{0.51\linewidth}
\vspace{0pt}
\caption{W2A2 quantization comparison.}
\label{tab:ablation}
\centering
\footnotesize
\setlength{\tabcolsep}{1.2pt}
\begin{tabular}{lccccc}
\toprule
Method & $S_\alpha\uparrow$ & $F_\beta^w\uparrow$ &
$E_\phi\uparrow$ & $F_\beta^m\uparrow$ & MAE$\downarrow$\\
\midrule
Baseline & .795 & .681 & .842 & .749 & .057\\
Channel-only & .799 & .688 & .848 & .749 & .055\\
Frequency-only & .802 & .694 & .854 & .758 & .054\\
\method{} no GGN & .813 & .719 & .875 & .774 & .050\\
\method{} & \textbf{.819} & \textbf{.727} & \textbf{.884} & \textbf{.780} & \textbf{.048}\\
\bottomrule
\end{tabular}
\end{minipage}
\end{table}

\begin{figure}[!t]
\centering
\includegraphics[width=\linewidth]{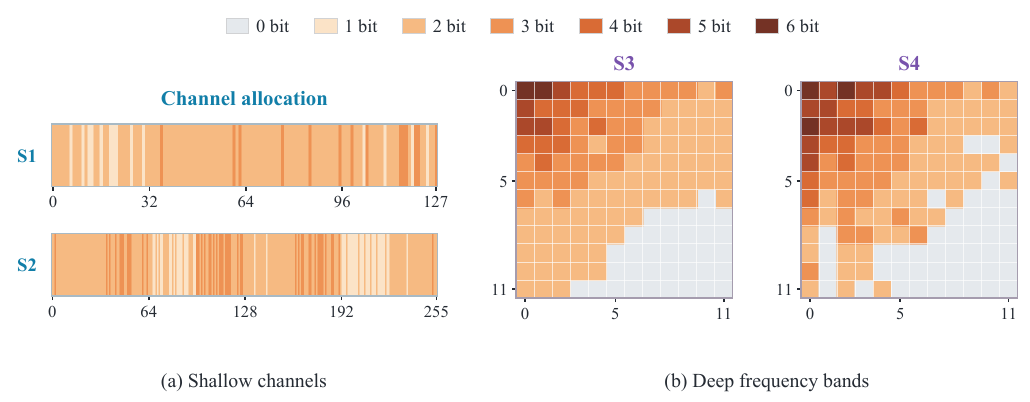}
\caption{Actual CFRN W2A2 bit assignments in the first attention-output
projection of each encoder stage. Channels retain their original order;
frequency maps use DCT coordinates. Every shown layer averages two bits.
Gray indicates suppressed, zero-bit bands.}
\label{fig:bit-allocations}
\end{figure}

The full configuration improves structural agreement and prediction alignment
while reducing pixel-wise error.
These gains suggest improvements in both the overall structure of the
predicted masks and their pixel-level accuracy.
The improvements over no-GGN favor the complete design.
Boundary-weighted reconstruction shows a similar pattern in the separate
CFRN W2A4 comparison on CHAMELEON, supporting the relevance of
boundary-focused reconstruction to overall mask quality.

Figure~\ref{fig:bit-allocations} visualizes CFRN's frozen W2A2 bit assignments.
Most shallow channels retain two bits, with selected channels adjusted to
one or three bits.
Deep layers show greater variation across frequency bands: lower frequencies
generally receive more precision, while part of the high-frequency region
is suppressed with zero-bit assignments.
Each displayed layer maintains an exact two-bit average.
The differences between the S3 and S4 maps reflect layer-specific calibration
costs, allowing each deep layer to distribute precision differently within
the same frequency domain.
This pattern illustrates how \method{} redistributes a limited precision
budget across different feature components, using channels in shallow stages
and local frequencies in deep stages.

\section{Conclusion}

We presented \method{}, a depth-adaptive post-training quantization framework
for camouflaged object detection.
Motivated by the changing characteristics of hierarchical features,
\method{} allocates activation precision across shallow channels and deep
local-frequency bands under a fixed per-layer average-bit budget.
Empirical-Fisher channel costs and COD-aware frequency costs guide this
allocation toward components important for foreground and boundary prediction.
We also formulated an auxiliary boundary-aware Hessian approximation that
incorporates boundary-weighted gradient and error statistics into reconstruction.
Experiments across COD models, benchmarks, and precision settings demonstrate
the effectiveness of the mixed-domain allocation.
This work highlights the importance of adapting both the allocation domain
and its sensitivity measure to preserve task-relevant information in low-bit
hierarchical COD models.

\subsection*{Ethics statement}

This work studies low-bit quantization for camouflaged object detection using
publicly available benchmarks. It involves no human-subject studies or
collection of new personal data. Although efficient COD can support
applications such as wildlife monitoring, its potential use in surveillance
and military applications warrants consideration.

\subsection*{Reproducibility statement}

We provide the implementation and configuration files for \method{}, along
with environment requirements and instructions for quantization and evaluation.
These materials document the calibration data, precision settings, and
reconstruction procedure, and include commands for evaluating the models on
all four COD benchmarks.

\subsection*{AI use statement}

We used generative AI tools to assist with code implementation and debugging,
experimental analysis, and manuscript preparation, including English editing
and LaTeX formatting. Reported results are based on experimental measurements
and cited sources. The authors take responsibility for the accuracy and
originality of the final manuscript and accompanying code.

\bibliography{m3_iclr2027}
\bibliographystyle{iclr2027_conference}

\clearpage
\appendix
\section{Expectation of the COD-GGN probe}
\label{app:ggn}

COD-GGN uses teacher-derived output weights to account for foreground--background
balance, boundaries, and prediction uncertainty during frequency allocation.
Here we explain how a random probe transfers these weights to layer
activations using a vector--Jacobian product (VJP) via first-order
backpropagation, without explicitly forming the Jacobian or Hessian.

For a fixed calibration image, let $\mJ_\ell$ be the Jacobian of the $H\times W$
output logits with respect to the flattened activation at layer $\ell$.
The model and the nonnegative teacher weights $\Omega$ in
Equation~\ref{eq:omega} are held fixed; the expectation is over independent
Rademacher entries of $\vepsilon$, each equally likely to be $-1$ or $+1$.
Thus $\mathbb{E}[\vepsilon]=0$ and
$\mathbb{E}[\vepsilon\vepsilon^\top]=\mI$. Defining
$\vv=\operatorname{diag}(\Omega)^{1/2}\vepsilon/\sqrt{HW}$ and
$\vg_\ell=\mJ_\ell^\top\vv$ gives
\begin{align}
\mathbb{E}[\vg_\ell\vg_\ell^\top]
&=\mJ_\ell^\top\mathbb{E}[\vv\vv^\top]\mJ_\ell\\
&=\frac{1}{HW}
\mJ_\ell^\top\operatorname{diag}(\Omega)\mJ_\ell .
\end{align}
Because $\Omega\ge0$, the expected matrix is positive semidefinite.
After expressing $\vg_\ell$ in the orthonormal block-DCT basis, its
component-wise squares estimate the diagonal curvature in that basis.
These sensitivities weight the coefficient errors in
Equation~\ref{eq:task-cost}, making errors in task-sensitive directions more
costly. The allocation cost uses this diagonal approximation and omits
cross-coordinate curvature interactions.

\section{Detailed gains over matched baselines}
\label{app:detailed}

Table~\ref{tab:detailed-deltas} expands the CFRN comparison in
Table~\ref{tab:precision} into per-dataset gains across five precision
settings. Each entry reports \method{} minus the precision-matched
baseline, exposing variation hidden by the macro averages. Gains are
positive for the first four metrics and negative for MAE.

\begin{table}[h]
\caption{Per-dataset metric changes of \method{} relative to the
precision-matched baseline, rounded to three decimal places.}
\label{tab:detailed-deltas}
\centering
\setlength{\tabcolsep}{3.0pt}
\begin{tabular*}{\linewidth}{@{\extracolsep{\fill}}llrrrrr@{}}
\toprule
Setting & Dataset & $\Delta S_\alpha$ & $\Delta F_\beta^w$ &
$\Delta E_\phi$ & $\Delta F_\beta^m$ & $\Delta$MAE\\
\midrule
W2A2 & CAMO & +.052 & +.082 & +.067 & +.062 & -.017\\
& CHAMELEON & +.025 & +.044 & +.051 & +.023 & -.009\\
& COD10K & +.011 & +.030 & +.026 & +.020 & -.003\\
& NC4K & +.010 & +.025 & +.023 & +.016 & -.005\\
\midrule
W2A3 & CAMO & +.017 & +.026 & +.019 & +.024 & -.005\\
& CHAMELEON & +.001 & +.001 & +.006 & .000 & .000\\
& COD10K & +.002 & +.004 & +.002 & +.004 & -.001\\
& NC4K & +.003 & +.004 & +.005 & +.004 & -.001\\
\midrule
W2A4 & CAMO & +.006 & +.009 & +.009 & +.009 & -.003\\
& CHAMELEON & .000 & +.002 & -.004 & +.004 & +.001\\
& COD10K & +.001 & +.003 & +.001 & +.005 & .000\\
& NC4K & .000 & +.001 & +.001 & +.001 & .000\\
\midrule
W3A3 & CAMO & +.016 & +.023 & +.019 & +.020 & -.005\\
& CHAMELEON & +.003 & +.007 & +.007 & +.004 & -.001\\
& COD10K & +.004 & +.008 & +.008 & +.006 & -.001\\
& NC4K & +.002 & +.004 & +.005 & +.002 & -.001\\
\midrule
W4A4 & CAMO & +.009 & +.017 & +.009 & +.014 & -.004\\
& CHAMELEON & +.003 & +.008 & +.005 & +.005 & -.002\\
& COD10K & +.010 & +.022 & +.010 & +.018 & -.003\\
& NC4K & +.008 & +.014 & +.007 & +.011 & -.003\\
\bottomrule
\end{tabular*}
\end{table}

Gains peak at W2A2 and generally shrink as activation precision rises with
two-bit weights, consistent with allocation being most useful under tight
budgets. Yet W4A4 gives larger weighted F-measure gains than W2A4 on every
dataset, showing that activation precision alone does not explain the
benefit. CAMO leads in four settings and COD10K at W4A4.

Weighted F-measure improves in all 20 dataset--precision pairs, while MAE
decreases in 16. These gains generally combine better foreground prediction
with lower pixelwise error, although the small MAE increase on CHAMELEON
at W2A4 shows that improvement is not uniform across metrics.

\section{Fixed hyperparameters and quantization scope}
\label{app:hyperparameters}

Table~\ref{tab:hyperparameters} summarizes the calibration, reconstruction,
and allocation settings shared by the main mixed-domain experiments.
The stated activation precision is an exact per-layer average over eligible
channels or frequency bands, allowing individual units to receive different
bit widths. The reconstruction component study follows the separate
protocol in Appendix~\ref{app:boundary-history}.

Fisher allocation covers eligible non-frequency Conv/Linear paths, including
the decoder; DCT allocation covers frequency-eligible stage 3--4 Linear
layers. Zero-bit bands are suppressed and patch-embedding activations
remain A8. Unwrapped operations follow the reconstruction baseline.

\begin{table}[H]
\caption{Configuration used for the main mixed-domain runs.}
\label{tab:hyperparameters}
\centering
\begin{tabular}{ll}
\toprule
Item & Value\\
\midrule
Input resolution & $384\times384$\\
Calibration / reconstruction images & 128 / 1,024\\
Fisher image--mask pairs / frequency images & 64 / 16\\
Reconstruction iterations & 20,000 per block\\
Reconstruction method & Least-squares Hessian + AdaRound + QDrop\\
QDrop probability / temperature & 0.5 / 20\\
Channel / DCT encoder stages & 1--2 / 3--4\\
Channel allocation spread $\gamma$ & 0.30\\
DCT block / bands & $12\times12$ / 144\\
Deep bit tiers & $\{0,2,3,4,5,6\}$\\
Frequency clip candidates & 8\\
COD-GGN blend $\beta$ & 0.10\\
Boundary kernels / weight & $\{3,7,15\}$ / 2.0\\
Uncertainty floor / curvature cap & 0.10 / 8.0\\
Frequency normalization reference & 2 bits for every precision\\
Patch embedding activation & 8 bits for every precision\\
\bottomrule
\end{tabular}
\end{table}

\section{Boundary-reconstruction component protocol}
\label{app:boundary-history}

Table~\ref{tab:boundary-reconstruction} uses a separate CFRN W2A4 protocol.
Both runs restore the same calibration checkpoint and use identical adaptive
iteration counts
across 35 blocks (153,800 iterations total).  Their shared teacher BCE uses
pixel weights $1+10|\operatorname{AvgPool}_9(\vp)-\vp|$, normalized by the
sum within each image.  Only the variant adds the local Hessian weighting
in Equation~\ref{eq:reconstruction-boundary-weight}, with
$\alpha_{\mathrm{rec}}=2$ and kernels 7/5/3 for patch embedding/stages 1/2.

\begin{table}[H]
\centering
\begin{minipage}[t]{0.49\linewidth}
\vspace{0pt}
\caption{Historical CFRN W2A4 reconstruction on CHAMELEON, with and
without local boundary weights.}
\label{tab:boundary-reconstruction}
\centering
\small
\setlength{\tabcolsep}{1.5pt}
\begin{tabular*}{\linewidth}{@{\extracolsep{\fill}}lccccc@{}}
\toprule
\shortstack[l]{Boundary\\weights} & $S_\alpha\uparrow$ & $F_\beta^w\uparrow$ &
$E_\phi\uparrow$ & $F_\beta^m\uparrow$ & MAE$\downarrow$\\
\midrule
Without & .897 & .841 & .942 & .878 & .027\\
With & .901 & .844 & .946 & .878 & .025\\
\bottomrule
\end{tabular*}
\end{minipage}\hfill
\begin{minipage}[t]{0.49\linewidth}
\vspace{0pt}
\caption{\method{} latency on CamoFormer-P under CUDA Graph replay
(RTX 5090). Speedup is relative to FP32.}
\label{tab:w4a4-latency}
\centering
\small
\setlength{\tabcolsep}{1.5pt}
\begin{tabular*}{\linewidth}{@{\extracolsep{\fill}}lrrr@{}}
\toprule
Implementation & \shortstack{Latency\\(ms)$\downarrow$} & FPS$\uparrow$ & Speedup$\uparrow$\\
\midrule
FP32 & 17.13 & 58.4 & 1.00$\times$ \\
W4A4 (integer) & 15.95 & 62.7 & 1.07$\times$ \\
W4A4 (FP16) & \textbf{11.28} & \textbf{88.6} & \textbf{1.52$\times$} \\
\bottomrule
\end{tabular*}
\end{minipage}
\end{table}

\section{W4A4 deployment performance}
\label{app:latency}

Deployment gains.
With CUDA Graph replay (Table~\ref{tab:w4a4-latency}), our fused W4A4
implementation runs CamoFormer-P at
88.6 FPS (11.28\,ms), achieving 3.33$\times$
and 1.52$\times$ speedups over the original PyTorch quantization
simulation and FP32 reference, respectively. These correspond to latency
reductions of 70.0\% and 34.1\%.
The packed-W4 integer-GEMM prototype reaches 62.7 FPS, a 2.36$\times$
speedup over the simulation. Both reuse the saved \method{} bit assignments
and quantization parameters without recalibration or reconstruction.

Accuracy after deployment.
On all 250 CAMO test images, the fused implementation achieves
$F_\beta^w=0.8164$ and $S_\alpha=0.8670$, compared with $0.8179$ and
$0.8679$ for the original quantized reference, respectively.
The packed-W4 integer-GEMM prototype achieves $F_\beta^w=0.8191$ and
$S_\alpha=0.8691$. Both implementations retain comparable segmentation
accuracy. Metrics come from fresh checkpoint replays; floating-point casts
and summation order cause small numerical differences.

Implementation and timing.
The fused path caches quantized weights and fuses quantization/dequantization
(Q/DQ), retaining DCT/IDCT and using FP16 GEMM. The packed-W4 path adapts
the COD-TDQ Triton backend~\citep{li2026codtdq}: weights are unpacked into
INT8 registers for Tensor Core products with INT32 accumulation.
It covers all 205 Linear layers and 76 pointwise convolutions, with FP16
outputs and remaining operators. A4 denotes the per-layer average activation
budget, including mixed tiers.

Measurements use PyTorch 2.8.0, CUDA 12.8, Triton 3.4.0, and driver 580.76.05
on an RTX 5090 (32\,GB), with cuDNN autotuning. All implementations use
batch size 1, $384\times384$ inputs, and disabled TF32; FPS is the
reciprocal of latency. We average five rounds of
100 complete forwards after 30 warm-up calls using synchronized CUDA events.
Timing includes channel gathering and padded computation, Q/DQ, DCT/IDCT,
the decoder, and all prediction heads. Inputs reside on the GPU; I/O,
preprocessing, postprocessing, compilation, graph capture, and offline
weight packing are excluded.

\end{document}